\documentclass[letterpaper, 10 pt, conference]{ieeeconf}  

\IEEEoverridecommandlockouts                              
\usepackage[utf8]{inputenc}
\usepackage[T1]{fontenc}
\usepackage{graphicx}
\usepackage{adjustbox}
\usepackage{amsmath}
\usepackage{comment}
\usepackage{makecell}
\usepackage{dblfloatfix}
\usepackage{hyperref}
\def\infinity{\rotatebox{90}{8}}

\title{\LARGE \bf
CIPCaD-Bench: Continuous Industrial Process datasets for benchmarking Causal Discovery methods
}

\author{Giovanni Menegozzo, Diego Dall'Alba and Paolo Fiorini{}
\thanks{*This work has been partially supported by the project "Dipartimenti di Eccellenza 2018/2022" funded by
the Italian Ministry of Education, Universities and Research (MIUR)}
\thanks{The authors are with the Department of Computer Science, University of Verona, Verona 37100, Italy. \textbf{Corresponding author:}
        {\tt\small giovanni.menegozzo@univr.it}} \thanks{
        {\tt\small }}%
}

\usepackage{url}
\begin{document}

\maketitle
\thispagestyle{empty}
\pagestyle{empty}

\begin{abstract}
Causal relationships are commonly examined in manufacturing processes to support faults investigations, perform interventions, and make strategic decisions. Industry 4.0 has made available an increasing amount of data that enable data-driven Causal Discovery (CD). Considering the growing number of recently proposed CD methods, it is necessary to introduce strict benchmarking procedures on publicly available datasets since they represent the foundation for a fair comparison and validation of different methods. This work introduces two novel public datasets for CD in continuous manufacturing processes. The first dataset employs the well-known Tennessee Eastman 
 simulator for fault detection and process control. The second dataset is extracted from an ultra-processed food manufacturing plant, and it includes a description of the plant, as well as multiple ground truths. These datasets are used to propose a benchmarking procedure based on different metrics and evaluated on a wide selection of CD algorithms. This work allows testing CD methods in realistic conditions enabling the selection of the most suitable method for specific target applications. The datasets are available at the following link: [https://github.com/giovanniMen]
\end{abstract}

\section{INTRODUCTION} \label{intro}

The understanding of industrial processes is largely based on finding cause-effect relationships among process variables and parameters, even though these relationships are often informally estimated based on expert knowledge. In the industrial context, there are numerous approaches to investigate the cause-effect relationships at both operational and managerial levels. Knowledge-based methods such as the 5Why or the Fishbone diagram \cite{fish} were extensively adopted by manufacturer even if they are exposed to modeler's bias. Causal relationship are commonly used to understand the cause of an error, make changes to the final product through specific interventions or estimate the effect of new policies on the market \cite{review}. Consequently, with a knowledge-based approach, the burden of finding causal relationships becomes increasingly onerous for practitioners.

The advent of Industry 4.0 has made it possible to acquire an increasing number of process parameters thanks to the extended sensorization of the process supported by an advanced communication, acquisition and data management infrastructure. Thanks to this information, it is possible to introduce data-driven approaches capable of identifying the relationships between the variables of the entire plant \cite{miocase}. 

Over the last decade, the widespread adoption of deep learning (DL) models had a significant impact on research due to their ability to recognize complex patterns on large-scale dataset. However, these architectures rely on mere statistical associations between variables and thus usually fail to describe the system's underlying structure \cite{loca}. Causal models extract the causal relationships between variables by overcoming association with a structural hierarchy. The advantage of inferring causal rather than statistical models is in expressing a deeper understanding of the system's behaviors. 
The Structural Causal Model (SCM) framework describes causal models by combining intervention on distributions, counterfactual thinking, graphical models and inference from data \cite{Peters}. SCM is not limited in recognizing patterns but allows to reason about the underlying structure beyond the observed correlation, such as simulating interventions with \textit{do-calculus} or assuming hypothetical changes in the environments with \textit{counterfactuals} \cite{Pearl}. Thus, several efforts have been made to integrate associative methods within SCM \cite{mioral}. In particular, a common goal is to recognize the causal relationships, also referred as Causal Discovery (CD), from data and build a causal model from the observed system's behavior. This process has been facilitated by the graphical representation of causal models and sharing data among the research community \cite{pearl2}. Under specific assumptions (described in section \ref{methodSection}) that are not straightforward to satisfy, CD based on graphical models can reconstruct the causal structure of the system without directly intervening on the system \cite{Peters}. Therefore, they enable CD from observational data without performing randomized control tests. 

As reported in a recent survey \cite{review}, there is an increasing number of works using CD methods in  manufacturing for: management, root cause analysis, causality as a facilitator, and conceptual work based on causal theory. The considerable adoption of knowledge-based methods in manufacturing and the increased proposal of new data-driven CD approaches mark an essential opportunity for manufacturing research. 
This opportunity requires the urgent introduction of benchmarking on public datasets, since they are crucial to compare and validate such methods and they are essential to facilitate CD research in industrial applications.

This work introduces CIPCaD-Bench, a Continuous Industrial Process CAusal Discovery Benchmark. The benchmark is released with two novel datasets extracted from manufacturing plants and a comparison protocol based on different metrics. The metrics are based on both matrix and graph format, for an objective assessment of CD methods' performance. 
This work allows a comparison of CD methods on real production data and it support the selection of the most suitable algorithm depending on the specific industrial application. Indeed, some characteristics (such as speed, precision or robustness) of CD algorithms can be preferred depending on the final application.
To provide solid starting foundation, we applied the proposed methodology to a wide set of recent CD methods available in literature.
 
The contributions of this work are the following:
\begin{itemize}

\item Release of two public datasets specifically designed for CD from continuous manufacturing plants: a novel dataset from ultra-processes food manufacturing production and an adaptation of the Tennesee Eastman process simulator.

\item A state of the art benchmark of causal discovery methods with multiple metrics that highlight algorithm's characteristics on the proposed datasets.
\end{itemize}

In section \ref{relatedSection}, we describe the available framework and datasets for CD and we better clarify the necessity of introducing public datasets covering the manufacturing processes domain.
In section \ref{methodSection}, we  describe the CD methods and the assumptions required to enable CD from observational data. Section \ref{DatasetSection} presents the proposed datasets together with the metrics an benchmarking procedure. Finally, in section \ref{resultSection} and \ref{discussionSection}, we present and discuss the obtained results to highlight the main differences among the considered CD algorithms.

\section{Related work} \label{relatedSection}
A deep knowledge of the real system under investigation is required to reconstruct its ground truth causal structure. This is probably the main limiting factor to the  diffusion of CD datasets on real applications \cite{lens}.
Most of the works are based on simulated datasets where causal relationships are enforced in the synthetic models used for data generation \cite{lens}.
Despite that, a limited number of benchmark datasets and challenges have been proposed in the literature for CD in different environments. A cause-effect challenge was released in 2013 for pairwise (i.e., only two variables) causal discovery \cite{pair}. The peculiarity of this dataset is the different distributions that the cause and the effect have for each pair. In 2016, the benchmark dataset CauseEffectPairs composed of 100 different cause-effect relationships from various real-world domains was released \cite{pair2}. The CauseMe dataset proposed as challenge in 2019 is focused on multivariate time series and includes some real-world examples on earth science systems \cite{runge}. Lastly, in \cite{lens}, a data generation procedure is proposed for time series that can violate the underling assumptions. The data generators allow to reproduce a large set of behaviours on synthetic time series, allowing CD benchmarking on data with specific characteristics. However, we are not aware of any specific benchmark targeting continuous industrial processes.

In general, there is a trade-off between synthetic data with reliable and reproducible ground truth and performance exploration on real-world application. Although causality is an intuitive concept for humans and can be associated with many real case scenarios, reconstructing a reliable ground truth remains highly complicated \cite{runge}. Especially in the case of multiple variables, it requires that no other external events are co-causing systems' behaviors and all required assumptions on data are satisfied. Synthetic datasets can  test the boundaries of CD methods, resulting in a more precise comparison. However in a specific real context, some cause may be more important than others. For example, the implications of assigning an erroneous causal relationship between two variables in a safety critical process (e.g., a process involving toxic or medical substances) may be severe. Therefore, it is desirable to use a domain as close as possible to the real one to evaluate the performance of CD methods. For all these reasons, it is necessary to propose benchmark datasets targeting the manufacturing domain. To the best of our knowledge, no other dataset provides a full-plant description and can be used for CD in continuous manufacturing processes. Aware of this trade-off between more reliable synthetic datasets and datasets closer to the domain application we propose two contributions, one based on a simulation that allows a more accurate comparison and one acquired from a real plant to allow an evaluation in realistic conditions.

Given the growing number of proposed CD algorithms and related challenges, several CD frameworks have emerged. DoWhy was firstly released in 2018 by Microsoft with a focus on causal inference \cite{dowhy}. For the time-series domain, the most comprehensive framework is Tigramite, proposed by Runge in 2019 \cite{runge}. For pairwise or multivariate CD, other four frameworks can be found in literature: Py-causal \cite{py}, Causal-learn \cite{causallearn}, CDT and g-castle \cite{cdt,gcas}. What differentiates these frameworks is the selections of methods included and  the programming languages. Py-causal is a Python wrapper around the Java-based Tetrad package, CDT wraps in Python methods written in R and Java code. Causal-learn is a recently released Python translation of the Tetrad package, and G-castle is a Python framework focusing on gradient-based methods. A small overlap exists between the available methods implemented in these frameworks, however, the main differences are other characteristics such as parallelization, integration, documentation.  

\section{Methods} \label{methodSection}
\subsection{Assumptions}
In this section, we briefly describes the required assumptions
to perform CD from data. As mentioned in section \ref{relatedSection}, to benchmark algorithm performance with different data assumptions, we recommend using synthetic datasets. It should be noted that with a strict interpretation, the theoretical assumptions for enabling CD, cannot be satisfied for a complex real-world system. For a detailed discussion on assumptions and their limitations, refer to \cite{Peters,loca,Beckers}.
A traditional way to discover causal relations is to use interventions or randomized experiments, which is in many cases too expensive, time-consuming, or even impossible \cite{reviewCD}. We focus our comparison on graph-based CD methods, as they allow a graphical interpretation of the results, suitable for application in analytical and decision-making industrial processes. 
The following assumptions relate causal relations to probability densities and enable CD from observational data \cite{miocase}:
\begin{itemize}
    \item \textbf{Causal Markov:} For causally sufficient sets of variables, all variables are independent of their non-effects conditional on their direct causes (from probability independence to graph independence).
    \item \textbf{Causal faithfulness:} For a causally sufficient set of variables $V$ in a population $P$, the population density $ \mathcal{P} (P(V))$ is faithful to the causal graph over $V$ for $P$ (from graph independence to probability independence).
\end{itemize}
These assumptions allow associating each graph to a unique joint distribution and vice versa. Some algorithms also require causal sufficiency (i.e., exogenous variables are not considered). 

\subsection{Causal discovery algorithms}
 Following \cite{reviewCD,DoYou,Spirtes2016}, we can group causal discovery algorithms in five categories: 
\begin{enumerate}

\item \textbf{Prediction based methods:} These models use outcome prediction to estimate the influence of the cause variables under the aforementioned assumption. One of the most representative methods in this class is Granger causality \cite{Granger}. 
  
 \item \textbf{Asymmetry methods:} Asymmetry methods test which nodes are more likely to be a cause or effect using asymmetry in the distributions. These asymmetries can be exploited in multiple ways. If some dependant noise in added to one variable, if this variable is a cause it will propagate to its effect variable. While if the variable is an effect, the system does not report any change to the cause variable.

\item \textbf{Score methods:} Score-based methods search over the space of possible graphs trying to maximize a score function that reflects the most suitable graph to fit the data. This score is typically related to the likelihood of the graph given the data. However, the number of possible graphs is super-exponential to variables.

\item \textbf{Gradient base methods:} Are similar to score based methods but they learn the causal structure as a continuous optimization problem, based on a smooth characterization of acyclicity \cite{gcas}. 
    
\item \textbf{Constraint methods:} Constraint based methods use independence and dependence constraints obtained from statistical tests to narrow down the candidate graphs that may have produced the data. They use independence tests to remove possible false causal relationships and orientation rules to create a subset of potential causal graph structures. 
\end{enumerate}

\subsection{Time series causal discovery}
Reasoning on causal relations among variables that refer to time series is generally easier than causal reasoning without time as the temporal structure entails additional constraints \cite{Peters}. However, while performing CD on time series additional requirements occurs. Causal discovery algorithms for time series have to account for additional concerns such as stationarity, instantaneous causal effect, subsampling and time delay between cause and effects \cite{reviewCD}. In this work, we focus on the independent and identically distributed (i.i.d.) setting to provide a more understandable and unified comparison. One of the two datasets provided, however, allows time series CD as the temporal ground truth is provided.

\subsection{Selected methods} \label{selectedMethods}
In Table \ref{tab:methodsComp} we list the algorithms considered for the benchmark. The methods have been chosen including a broad selection of approaches while maintaining a uniform implementation. The goal is to analyse the CD's performance of the methods with various metrics and facilitate the comparison with the benchmark. For this reason, we chose methods available in Python that belong to the presented categories. We excluded prediction-based methods because they were restricted to pairwise analysis. Following, we briefly describe the methods adopted in the comparison:
\begin{itemize}
    \item \textbf{ICA-LiNGAM:} It evolves from the LiNGAM method which uses the non-Gaussian structure of data to estimate the causal directions of variables. The Indipendent Component Analysis (ICA) method allows to improve the computation and scalability compared to standard LiNGAM. It requires the same interpretation of the assumptions of the LiNGAM algortim: no latent confounding variables,  non-Gaussianity, faithfulness and causal Markov. The algorithm return a Direct Acyclic Graph (DAG).
    \item \textbf{Direct-LiNGAM:} It guarantees to converge to a fixed number of steps differently from the ICA-LiNGAM. Similarly to the latter it requires the same strict models assumptions and is hard to scale. For an in depth comparison between LiNGAM, ICA-LiNGAM and Direct-LiNGAM refer to \cite{lingamcomp}. The algorithm return a DAG.
    \item \textbf{GES:} Greedy Equivalent Search is one of the most used score-based algorithms \cite{Ges}; instead of exploring the optimal DAG, it chooses a node and analyses possible neighbors. Then, it keeps adding dependencies between nodes until it reaches a maximum for each node. In the second step, it removes dependencies and produces an equivalence DAG (i.e., a set of possible causal graphs). The output is a Partially Directed Acyclic Graph (PDAG) that represent the Markov equivalence class. The only assumptions is linearity in the data. 
    \item \textbf{PC:} Peter Clark (PC) algorithm provides a search architecture based on statistical procedures \cite{PC}. The PC algorithm guarantees the convergence to the valid Markov equivalence class. PC returns a DAG and assumes causal sufficiency.
    \item \textbf{FCI:} Fast Causal Inference (FCI) algorithm has been proposed as an evolution of PC to allow the presence of unobserved variables and thus FCI does not assume causal sufficiency \cite{FCI}. It returns a maximal ancestral graph (MAG) instead of a DAG that uses bidirectional edges. Compared to PC, FCI optimizes speed and assumption requirements.
     \item \textbf{NOTEARS:}  NOTEARS is a method that aims to identify a DAG from the data that explains the residual variance. Even if in practice it has been proved ineffective for causal discovery as recently reported in \cite{causalens}, we include this method as it was one of the most common methods initially used.
      \item \textbf{NOTEARS-MLP:}  NOTEARS-MLP extend the DAG search alghoritm NOTEARS to non linear cases thanks to a Multi-Layer Perceptron (a Neural Network with more layers). However it suffers from the same problem of NOTEARS which consists on being affected by the different scales in the data. This may involve possible causal interpretations of associative relationships. It returns a DAG.
      \item \textbf{MCSL:}  It uses gradient-based optimization, by leveraging a smooth characterization of acyclicity and the Gumbel-Softmax approach to approximate the binary adjacency matrix super-graphs of the true causal graph. It return a DAG under the same assumptions of NOTEARS-MLP.
      \item \textbf{GOLEM:} It propose a likelihood-based structure learning  with continuous unconstrained optimization. It scales to a large number of variables and returns a DAG. 
     \item \textbf{CORL:} Incorporates reinforcement learning (RL) into the ordering based paradigm. A generated ordering can be pruned by variable selection to obtain the causal DAG. It assumes causal minimality and it returns a DAG.

\end{itemize}

\begin{table}[]
\begin{adjustbox}{width=\columnwidth}
\begin{tabular}{|c|c|c|c|c|}
\hline
Method& Description& Type& From& Implementation \\\hline
Direct-LiNGAM & \makecell{Linear Non-Gaussian Acyclic Model}& Asymmetry & \cite{DirectLingam} & causal-learn\\\hline
ICA-LiNGAM  &\makecell{Independent Component Analysis \\Linear Non-Gaussian Acyclic Model} & Asymmetry & \cite{NonLinLingam} & gCastle      \\ \hline
GES& Greedy Equivalence Search& Score     & \cite{Ges} & causal-learn   \\ \hline
PC& \makecell{Peters Clark alghoritm on \\conditional independence tests}& Constraint & \cite{PC} & causal-learn \\\hline
FCI& Fast Causal Inference& Constraint & \cite{FCI} & causal-learn   \\ \hline
NOTEARS& \makecell{A gradient-based algorithm for\\ linear data models} & Gradient  & \cite{noTears} & gCastle        \\\hline
NOTEARS-MLP & \makecell{NoTears  using neural network  \\ modeling for non-linear \\causal relationships  }  & Gradient  & \cite{NoTearsMLP} & gCastle      \\\hline
GOLEM& \makecell{Efficient version of NOTEARS} & Gradient  & \cite{Golem} & gCastle      \\\hline
CORL& \makecell{A reinforcement learning and\\ order-based algorithm}& Gradient  & \cite{CORL} & gCastle       \\\hline
MCSL& \makecell{Non-linear additive noise\\ data by learning the binary\\ adjacency matrix}           & Gradient  & \cite{MCSL} & gCastle \\\hline    
\end{tabular}%
\end{adjustbox}
\caption{Methods used for comparison in the CIPCaD-Bench. For each method, the type and frameworks are reported. }
\label{tab:methodsComp}
\end{table}
\section{Datasets and Metrics} \label{DatasetSection}
\subsection{\textbf{Tennessee-Eastman dataset}}
\subsubsection*{\textit{Overview}} 
\begin{figure}[h]
    \centering
    \includegraphics[width=0.5 \textwidth]{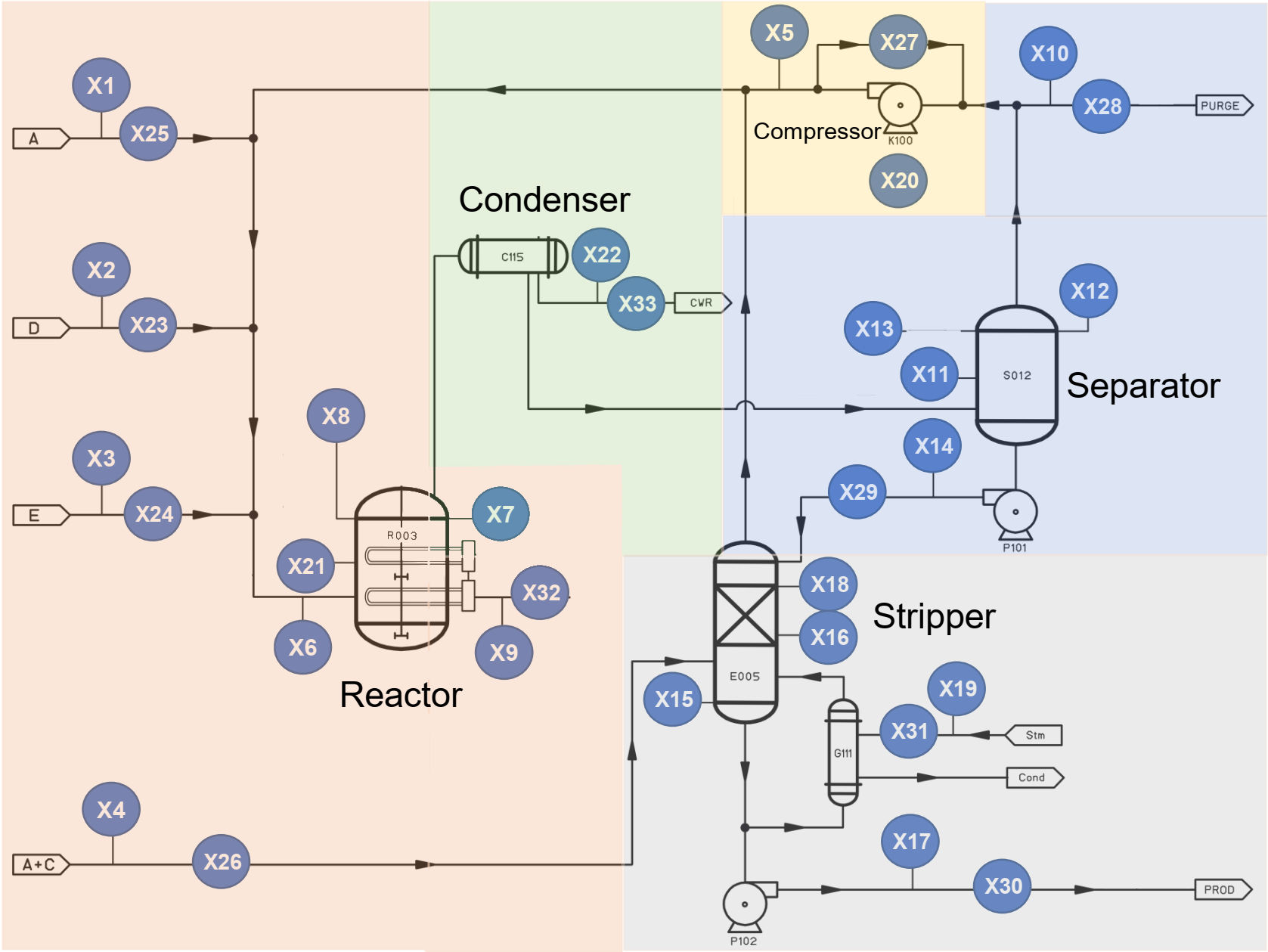}
    \caption{Structure of the TE production plant with corresponding variables selected for causal discovery. The colors represent the different process phases described in \cite{TECausalGT}}
    \label{fig:TeProcess}
\end{figure}

The Tennessee Eastman (TE) Process is a frequently used benchmark in chemical engineering research \cite{reviewTE}. It was firstly presented in 1992 by Downs and Vogel in \cite{TEOr}. Over time it has undergone several elaborations to fix the generative process, update the source code written (initially in Fortran) or to adapt to specific domain problems (e.g., fault detection, predictive monitoring, process control and so on \cite{TEext1,TEext2,TEext3,TEext4,TEext5,TEext6}). This process simulator is suitable for both process monitoring and root cause identification research due to its ability to simulate faults and the available description of the whole production process. The plant consists of five main units: a two-phase reactor, a condenser, a recycle compressor, a liquid-vapor separator and a product stripper. The process involves 41 measured variables and 12 manipulated variables. The measured variables include heterogeneous measurements that can be monitored continuously or sampled over a period of time, such as: pressures, levels, temperatures, concentrations. The sampling rate for all measured variables was set to 0.1 Hz. The plant is represented in Figure \ref{fig:TeProcess}. For in depth description of the TE refer to \cite{TEext2}.

\subsubsection*{\textit{Preprocessing}}
We elaborate the data generated for the normal steady state from \cite{TEext2}. The data is subsampled using a median operator on a 3 minutes sliding window over a total of 75 hours. As in \cite{TECausalGT}, we selected a total of 33 variables reported in Table \ref{TEvar}. They are divided in $22$ consecutive measurements $(x1 -x22)$ and $11$ manipulated variables $(x23 -x33 )$. We removed the variables with null variance. An analysis of the distributions of the variables is given in the supplementary materials.
\begin{table}[h]
\begin{adjustbox}{width=\columnwidth,center}
\begin{tabular}{|c|c|l|c|}
\hline
Name & Name in \cite{TEext2} & Description                                & Unit                     \\
\hline
X1   & XMEAS(1)                         & A Feed (stream 1)                          & kscmh  \\
X2   & XMEAS(2)                         & D Feed (stream 2)                          & kg/hr  \\
X3   & XMEAS(3)                         & E Feed (stream 3)                          & kg/hr  \\
X4   & XMEAS(4)                         & A and C Feed (stream 4)                    & kscmh  \\
X5   & XMEAS(5)                         & Recycle Flow (stream 8)                    & kscmh  \\
X6   & XMEAS(6)                         & Reactor Feed Rate (stream 6)               & kscmh   \\
X7   & XMEAS(7)                         & Reactor Pressure                           & kPa gauge\\ 
X8   & XMEAS(8)                         & Reactor Level                              & \%  \\
X9   & XMEAS(9)                         & Reactor Temperature                        & Deg C  \\
X10  & XMEAS(10)                        & Purge Rate (stream 9)                      & kscmh \\
X11  & XMEAS(11)                        & Product Sep Temp                           & Deg C \\
X12  & XMEAS(12)                        & Product Sep Level                          & \%   \\
X13  & XMEAS(13)                        & Prod Sep Pressure                          & kPa gauge\\ 
X14  & XMEAS(14)                        & Prod Sep Underflow (stream 10)             & m3/hr  \\  
X15  & XMEAS(15)                        & Stripper Level                             & \%   \\
X16  & XMEAS(16)                        & Stripper Pressure                          & kPa gauge \\ 
X17  & XMEAS(17)                        & Stripper Underflow (stream 11)             & m3/hr  \\
X18  & XMEAS(18)                        & Stripper Temperature                       & Deg C  \\
X19  & XMEAS(19)                        & Stripper Steam Flow                        & kg/hr  \\
X20  & XMEAS(20)                        & Compressor Work                            & kW      \\
X21  & XMEAS(21)                        & Reactor Cooling Water Outlet   Temp        & Deg C   \\
X22  & XMEAS(22)                        & Separator Cooling Water Outlet   Temp      & Deg C  \\  \hline
X23  & XMV(1)                           & D Feed Flow (stream 2)                     & \%  \\
X24  & XMV(2)                           & E Feed Flow (stream 3)                     & \% \\
X25  & XMV(3)                           & A Feed Flow (stream 1)                     & \% \\
X26  & XMV(4)                           & A and C Feed Flow (stream 4)               & \%  \\
X27  & XMV(5)                           & Compressor Recycle Valve                   & \%  \\
X28  & XMV(6)                           & Purge Valve (stream 9)                     & \% \\
X29  & XMV(7)                           & Separator Pot Liquid Flow   (stream 10)    & \% \\
X30  & XMV(8)                           & Stripper Liquid Product Flow   (stream 11) & \%  \\
X31  & XMV(9)                           & Stripper Steam Valve                       & \%  \\
X32  & XMV(10)                          & Reactor Cooling Water Flow                 & \%  \\
X33  & XMV(11)                          & Condenser Cooling Water Flow               & \%    \\   
\hline

\end{tabular}
\end{adjustbox}
\caption{TE variables description used for causal discovery. The horizontal line separate process variables (top) from manipulated variables (bottom). Refer to \cite{TEext2} for more details.}
\label{TEvar}
\end{table}
\subsubsection*{\textit{Causal Structure}}
\begin{figure}[h]
    \centering
    \includegraphics[width=0.5 \textwidth]{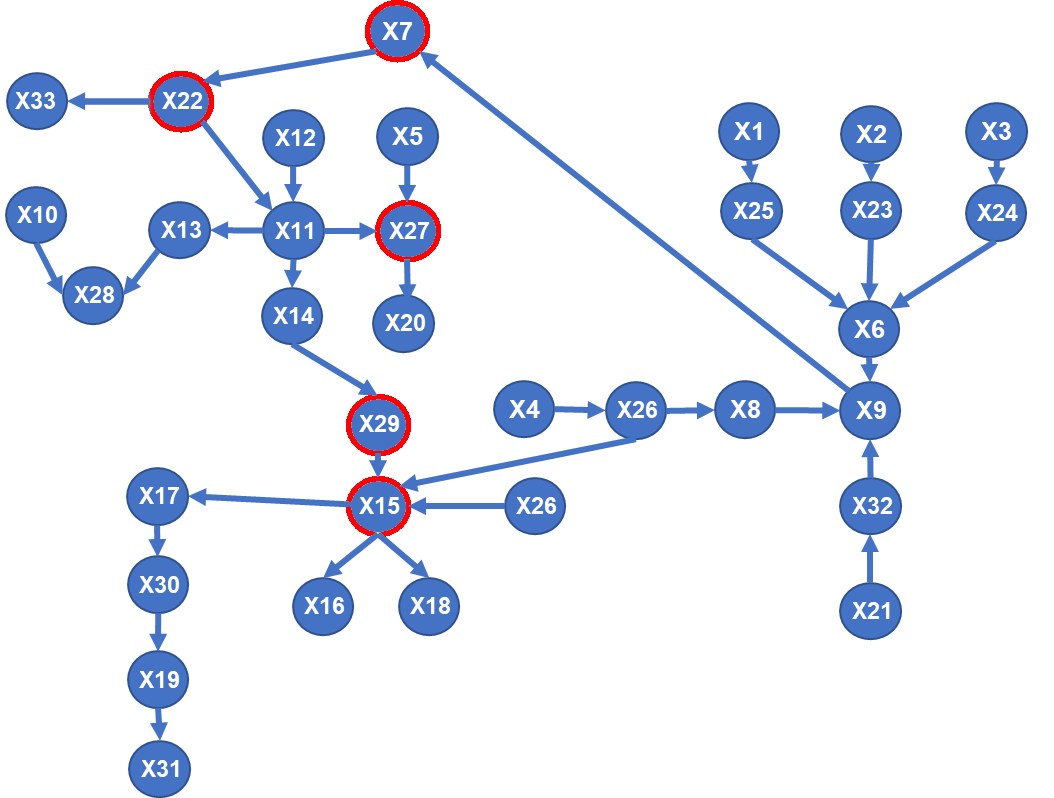}
    \caption{Ground truth causal graph for TE dataset. The ground truth was generated following the mechanical relationships of the plant as reported in \cite{TECausalGT}.}
    \label{fig:TeCausalStructure}
\end{figure}

According to the plant structure, the process is divided into five phases. Under the constraints of mechanical knowledge, the causal structure can be described as in \cite{TECausalGT}. Since TE dataset is constructed through simulation, it is possible to generate a ground truth for each variable. In the selected data, there are no automatically generated faults, therefore, it was not possible to determine the delay between the various cause-effect relationships. The order of the variables in the submodules is defined from their position in the productive process. The ground truth causal structure is reported in Figure \ref{fig:TeCausalStructure}.

\subsection{\textbf{Ultra-processed Food dataset}}
\subsubsection*{\textit{Overview}}
\begin{figure}[!h]
    \centering
    \includegraphics[width=0.4 \textwidth]{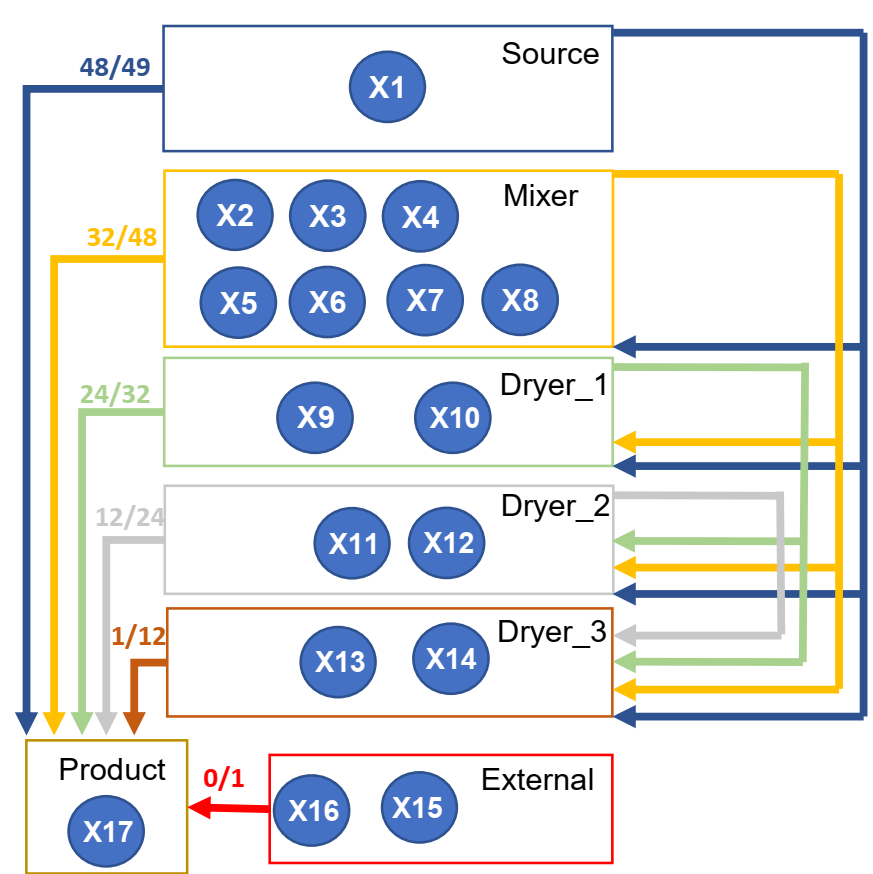}
    \caption{Description of the UF production flow. Variables are highlighted with circles while the temporal delay (in number of instances) is reported for each causal link. For example an intervention on variable $X2$ will have an effect on variable $X17$ with a delay between $32$ and $48$ temporal instances (i.e., 160 and 240 minutes).}
    \label{fig:UFProcess}
\end{figure}

Ultra-processed food (UF) companies are specialized in reducing natural raw material’s variations and provide customers products with more homogeneous characteristics. The raw data are obtained from an industrial plant for ultra-processed food production. The sampling was carried out every 5 minutes while the total production cycle takes approximately 3 hours, from raw ingredients to final semi-finished products. The extracted data represent approximately 80 days of production divided between a minimum of two and a maximum of four days of continuous recording per week. The process flow is shown in Figure \ref{fig:UFProcess}. Differently from the TE dataset, it is difficult to describe the process because of: complex dynamics of real time production which may be influenced by external causes, and for the confidentiality restrictions imposed by companies to avoid disclosing critical information on their production processes. Variables $2-14$ belonging to $4$ specific phases of the process and influence the qualitative variable $17$. Variables $15$ and $16$ are external variables not controlled by the process which affect the final product.  It should also be noted that some variation may be due to changes in raw materials, in production flow (variable $1$) or to possible reconfiguration between weeks. However while the magnitude of effects may change between weeks, the causal relationships are dictated by the plant and process dynamics and are consistent (at the best of potential un-cofounder and faults) throughout the production.
\subsubsection*{\textit{Preprocessing}}
We normalized and anonymized the dataset. We selected 17 variables belonging to different production machines. We applied a median operator on a 30 minutes moving window  to reduce the influence of improbable values. Data are standardized using a robust scaler that removes the median and scales the data according to the quartile range. We removed the variables with null variance. An analysis of the distributions of the variables is given in the supplementary materials.
\subsubsection*{\textit{Causal Structure}}
Since the Ultra-processed Food (UF) dataset is extracted from a running production, it is not possible to trace all the relations of the plant due to uncontrolled events or uncertainties. However, it is possible to establish an order between the machines and consequently to set up a production flow. The UF dataset is particularly  suited to revealing false negatives (i.e., causal links that are identified but violate the process flow).
As for the causal relationships between the parameters of the same machine, it is not possible to establish with certainty the causal direction. To compensate for this missing information, two ground truths have been proposed. The difference between the two causal models 
is in the relationships between the variables of the same machine. The UF ground truth does not include all the relationships belonging to the same machine, while the Ultra processed Food with Internal Machine Dependencies (UFIMD) fully connects each parameters of the same machine. The parameters belonging to different machines are represented inside the same box in Figure \ref{fig:UFProcess}. Consequently, the false positives for the UFIMD ground truth will be exceptionally reliable because each false positive represents a violation of the process flow. UF ground truth instead, considers the precision of the methods in identifying more relevant causal relationships between the different machines.
Finally, a summary of the different features for TE and UF dataset is reported in Table \ref{tab:DatasetProp}. Overall, the datasets describe a total of $223$ causal relationships between variables with the TE being lowest with $32$ causal relationships while the UF and the UFIMD entails $83$ and $108$ relationships respectively.
\begin{table}[h]
\begin{adjustbox}{center}
\begin{tabular}{|l|l|l|}
\hline
Features    & TE      & UF      \\\hline 
Frequencies    & 3 min   & 5 min   \\
N\_Samples     & 1500    & 23132   \\
N\_Variables   & 33      & 17      \\
Process Lenght & Unknown & 4 hours \\
Un-cofounders   & No      & Yes     \\
Sufficiency     & Yes     & No      \\
Outliers       & No      & Yes    \\\hline
\end{tabular}%
\end{adjustbox}
\caption{Proprieties for Tennesee Eastman (TE) and Ultra-processed food continuous process (UF) datasets.}
\label{tab:DatasetProp}
\end{table}
\begin{table}[]
\begin{adjustbox}{width=\columnwidth,center}
\begin{tabular}{|l|l|l|l|l|}
\hline
Name & Acr & Description& Best & Worst \\
\hline
True positive&TP& Detected with correct direction&+\infinity&0\\ 
True negative&TN& \makecell[l]{Neither in estimated\\ nor in true graph}&+\infinity&0\\
False positive&FP& \makecell[l]{Estimated but\\ not present in the true graph}&0&+\infinity\\ 
False negative&FN& \makecell[l]{Missed true causation \\in the estimated graph}&0&+\infinity\\ 
Reverse&R& \makecell[l]{Reversed direction in the\\ estimated graph}&0&+\infinity\\ 
False discovery rate &FDR& Rate for false discovered &0&1\\ 
True positive rate &TPR& Rate for true discovered&1&0\\ 
False positive rate&FPR& \makecell[l]{Rate for false detection}&0&+\infinity\\ 
\makecell[l]{Structural Hamming\\ Distance}&SHD&\makecell[l]{Number of operation (add/ \\ remove/flips) for estimated graph}&0&+\infinity\\
Precision&PR& \makecell[l]{Ratio of correct links \\with respect to FP}&1&0 \\ 
Recall&RE& Ratio of correct links detected&1&0\\ 
Undirected missing &UM& FN for undirected  link&0&+\infinity\\
Undirected extra &UE& FP for undirected  link&0&+\infinity \\
F1-score&F1& Is a measure of a test's accuracy&1&0\\ 

\hline
\end{tabular}
\end{adjustbox}
\caption{Description of the metrics used as comparison}
\label{BasicMetrics}
\end{table}
\subsection{Metrics} \label{sectionMetrics}
As reported in the introduction, to evaluate a causal model different metrics can be used depending on the aim of the final application. We provide several metrics for model evaluation summarized in Table \ref{BasicMetrics}. The equations reported in (\ref{formula}) define the following metrics:
\begin{align} \label{formula}
 FDR &= \frac{R + FP}{TP + FP}\\ \notag
 TPR &= \frac{TP}{TP + FN}	  \\ \notag
 FPR &= \frac{R + FP}{TN + FP}	  \\ \notag
 SHD &= UE + UM + R	  \\ \notag
 PR &= \frac{TP}{TP + FP}	  \\ \notag
 RE &= \frac{TP}{TP + FN}\\	  \notag
 F1 &= \frac{2*RE*PR}{RE+PR} \\ \notag
\end{align} 
\subsection{Comparison} 
Since the methods considered for the comparison have different outputs and they are working under diverse assumptions (see Section \ref{selectedMethods}), we must introduce some approximations to obtain uniform and comparable results:
\begin{itemize}
    \item We decompose each bidirectional edge into two edges with opposite direction such as $A \leftrightarrow B$ $=$ $A \leftarrow B$ , $A \rightarrow B$.
    \item Edges arising from un-cofounded variables (as in the case of the FCI algorithm) were approximated as directed edges.
    \item Non-directed edges are reported as missing edges since the method cannot distinguish the correct causal direction. 
\end{itemize}
We used graph-based and matrix-based metrics to allow a complete evaluation of the methods, considering different characteristics, from the control on false-positives to the ability of estimating the edges directionality.
\begin{figure*}[h]
    \centering
    \includegraphics[width=10.5 cm]{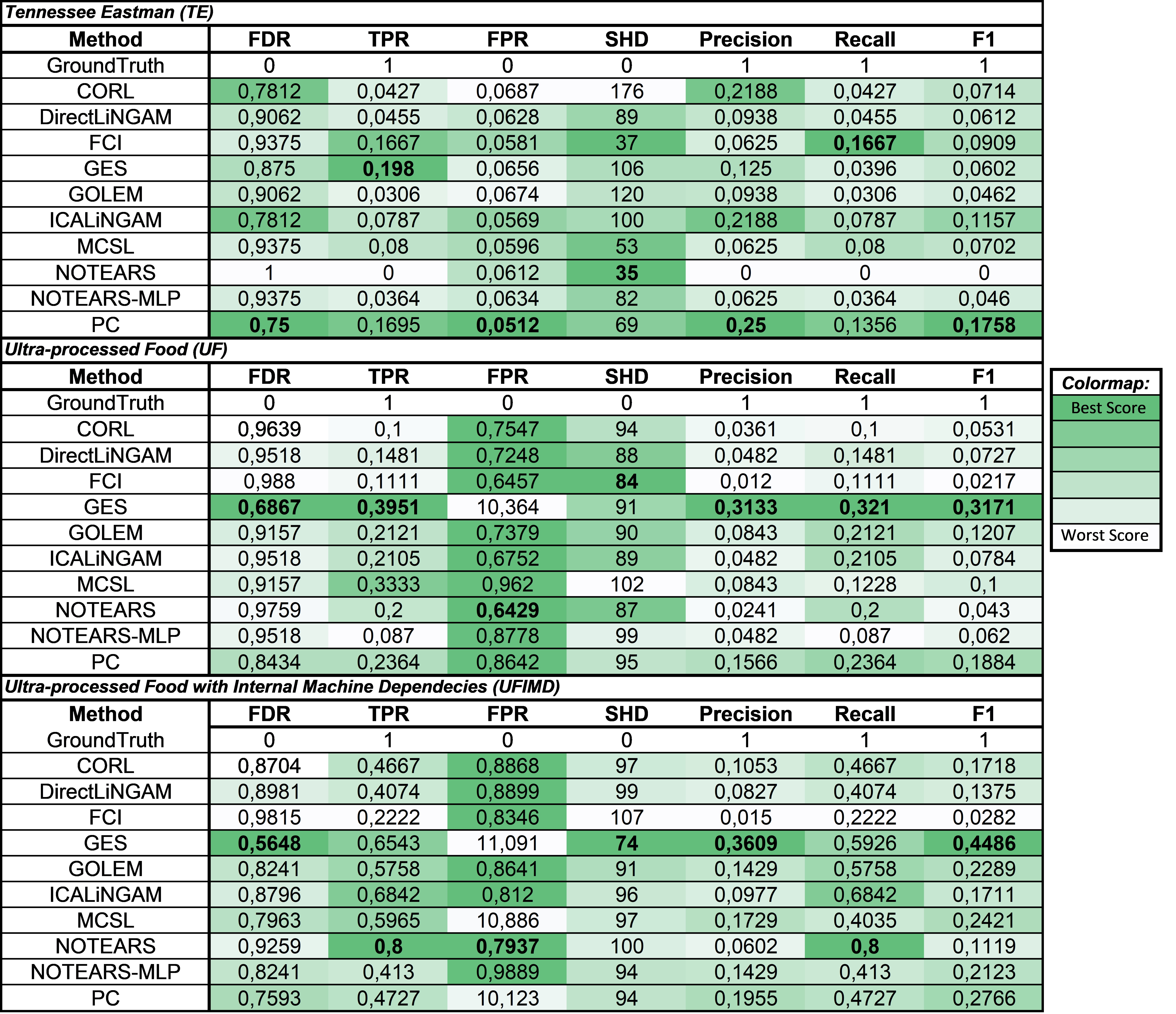}
    \caption{Results of the comparison methods. The first line for each dataset report the score of the respective ground truth. Best scores for each metrics are highlighted in bold. The color-scale indicate the intensity within each metrics for each dataset.}
    \label{tab:ResultsComparison}
\end{figure*}

\section{Results} \label{resultSection}
In Figure \ref{tab:ResultsComparison} we report the results of the experiments with the metrics presented in \ref{sectionMetrics} on the TE, UF and UFIMD datasets. In the results, we highlight the best results for each dataset and for each metric. For ease of reading, we highlight the results using a colormap. 
In Table \ref{tab:BestScore} we report the best method for each metric. This comparison supports the discussion of what algorithms are the most rigorous or explorative.

\section{Discussion} \label{discussionSection}
 One of the first aspects that emerge from the comparison between the three datasets is related to the impact of the number of causal relationships on the ground truth. This difference is particularly evident when directly comparing results obtained on UF and UFIMD datasets for the same methods. From Figure \ref{tab:ResultsComparison} it can be elaborated that the average TPR for the UFIMD dataset $(0.53)$ is more than twice the average TPR of the UF dataset $(0.20)$ clarifying that the number of causal relationships in the ground truth affects the absolute value of the TPR metric.
 The UF dataset and even more the UFIMD provide a ground-truth with many causal relationships, therefore, the number of missed dependencies (false negatives) is comparable to the number of false positives (i.e., finding a causal link that does not exist). This is less convenient on the TE dataset that contains few causal relationships with respect to the number of variables, and presents large differences between FPR (in average 0.06) and FDR (in average  0,88) metrics. From these considerations it can be deduced that UF and UFIMD datasets are more suited to investigate the precision metric, because they enable the analysis of wrong predictions that reverse the direction of the production flow. On the other hand, the TE dataset is more suitable to evaluate the ability of CD methods to limit the number of undetected causal links (i.e., the recall metric).
 \begin{table}[!htp]
\begin{adjustbox}{center}
\begin{tabular}{|l|l|l|l|}
\hline
\textbf{Metric} &\textbf{TE}      & \textbf{UF} & \textbf{UFIMD}     \\\hline 
FDR    & \textit{PC}   & \textit{GES} & \textit{GES}   \\
TPR     & \textit{GES}    & \textit{GES} & \textit{NOTEARS}   \\
FPR   & \textit{PC}      & \textit{NOTEARS} & \textit{NOTEARS}      \\
SHD & \textit{NOTEARS} & \textit{FCI} & \textit{GES} \\
Precision   & \textit{PC}      & \textit{GES} & \textit{GES}     \\
Recall     & \textit{FCI}     & \textit{GES} & \textit{NOTEARS}      \\
F1       & \textit{PC}      & \textit{GES} & \textit{GES}    \\\hline
\end{tabular}%
\end{adjustbox}
\caption{Best method by metrics and datasets}
\label{tab:BestScore}
\end{table}
 In causal discovery, a trade-off often occurs between setting CD method's parameters to find many causal relationships with lots of false positives or fewer relationships with a high missing rate. Therefore, we can state that no method is unquestionably better than the others. 
 The relative performance between algorithms varies depending on the dataset considered. However, this work provides an objective methodology to evaluate multiple methods with different metrics and therefore choose the leading solution for a specific application. In Table \ref{tab:BestScore} we report the best CD method for each metric considered on the 3 different datasets; for example GES works very well in finding many causal relations but has higher false positive rates. 
 ICA-Lingam and PC present low false negatives in TE and UF; however, the latter fails to control the internal machine dependencies showing a possible limitation of this method in complex contexts.  In the TE dataset FCI gives the highest performance in terms of recall.
In terms of reliability, the best methods are PC and GES, which also present a high recall and F1.
To summarize, depending on the specific target application, the following conclusions can be drawn:
\begin{itemize}
\item If detecting causal relationships when they are not present in the real process is particularly harmful for the application (as in the case of plant flow's reconstruction or inline interventions), the PC algorithm can be considered the most suitable due to its high precision.
\item If missing causal relationships is particularly harmful (as in the case of features filtering or fault identification), the use of FCI seems more suitable due to its ability to control recall.
\item Finally, if there is no particular propensity towards one of the two types of error, the obtained results indicate that GES and PC methods are the best algorithms. 
\end{itemize}
The choice of the algorithm is subordinate to the assumptions made on the data and is bounded to other features such as cost of computation, facility of implementation, true positive and false negative ratio, and presence of outlier. 

\section{Conclusion}
In this paper, we presented CIPCaD-Bench, a Continuous Industrial  Process  CAusal  Discovery  Benchmark. Two datasets extracted from manufacturing continuous processes are relased. The datasets are available at the following link: [https://github.com/giovanniMen].
We consider a wide selection of algorithms and metrics for evaluation. In the discussion we suggest a guideline for using the correct causal discovery algorithm depending on the final application. As future work, we plan to extend this research to time series CD algorithms, since the UF dataset also gives temporal delay between cause and effect.

\bibliographystyle{IEEEtran}
\bibliography{IEEEabrv,IEEEexample}

\end{document}